\documentclass[11pt]{article}
\usepackage[margin=1in]{geometry}
\usepackage{marginnote}
\usepackage{authblk}

\usepackage{setspace}

\usepackage{amsmath}
\usepackage{mathtools}
\usepackage[numbers]{natbib}
\usepackage{bm,color}
\usepackage{xr}
\usepackage{amssymb}
\usepackage{multirow}
\usepackage{booktabs}
\usepackage{times,caption}
\usepackage[normalem]{ulem}
\usepackage[flushleft]{threeparttable}
\usepackage[figuresright]{rotating}
\usepackage{url}
\usepackage[utf8]{inputenc}
\usepackage{amsthm}
\usepackage{accents}
\PassOptionsToPackage{hyphens}{url}\usepackage[colorlinks=true,linkcolor=blue,citecolor=blue,urlcolor=blue]{hyperref}
\usepackage[font=small,labelfont=bf]{caption}
\usepackage{indentfirst}
\usepackage{bbm}

\newcommand{\mb}[1]{\mathbf{#1}}
\DeclareMathOperator{\E}{\mathbb{E}}

\title{Generalized Probabilistic Canonical Correlation Analysis for Multi-modal Data Integration with Full or Partial Observations}
\author[1]{Tianjian Yang}
\author[1,*]{Wei Vivian Li}
\affil[1]{Department of Statistics, University of California, Riverside}
\affil[*]{To whom correspondence should be addressed: weil@ucr.edu}
\date{}
\begin{document}
\maketitle

\begin{abstract}
\noindent\textbf{Background:} The integration and analysis of multi-modal data are increasingly essential across various domains including bioinformatics. As the volume and complexity of such data grow, there is a pressing need for computational models that not only integrate diverse modalities but also leverage their complementary information to improve clustering accuracy and insights, especially when dealing with partial observations with missing data.\\
\noindent\textbf{Results:} We propose Generalized Probabilistic Canonical Correlation Analysis (GPCCA), an unsupervised method for the integration and joint dimensionality reduction of multi-modal data. GPCCA addresses key challenges in multi-modal data analysis by handling missing values within the model, enabling the integration of more than two modalities, and identifying informative features while accounting for correlations within individual modalities. The model demonstrates robustness to various missing data patterns and provides low-dimensional embeddings that facilitate downstream clustering and analysis. In a range of simulation settings, GPCCA outperforms existing methods in capturing essential patterns across modalities. Additionally, we demonstrate its applicability to multi-omics data from TCGA cancer datasets and a multi-view image dataset.\\
\noindent\textbf{Conclusion}: GPCCA offers a useful framework for multi-modal data integration, effectively handling missing data and providing informative low-dimensional embeddings. Its performance across cancer genomics and multi-view image data highlights its robustness and potential for broad application. To make the method accessible to the wider research community, we have released an R package, GPCCA, which is available at \url{https://github.com/Kaversoniano/GPCCA}.
%
\end{abstract}

\section{Introduction} \label{sec:intro}

Many real-world datasets can be described from multiple perspectives, where each perspective, typically represented as a matrix, corresponds to a data modality. A dataset that is consisted of multiple modalities collected from the same set of individuals is termed a multi-modal dataset \cite{guo2019canonical}. Examples include medical imaging data combining computed tomography (CT) and magnetic resonance imaging (MRI) scans \cite{kalamkar2023multimodal} and multi-omics datasets from The Cancer Genome Atlas (TCGA) \cite{cancer2008comprehensive}, which include DNA methylation, messenger RNA (mRNA) expression, and microRNA (miRNA) expression. Technological advances have made the collection of multi-modal data increasingly prevalent, enabling integrative analyses that leverage information across modalities. Analyzing a single modality in isolation is often suboptimal, as each modality may provide distinct yet complementary signals, which, when integrated, can enhance learning performance \cite{hao2021integrated}.

A key challenge in multi-modal learning is to capture shared structures across modalities while leveraging modality-specific signals to improve inference and prediction. In the context of unlabeled multi-modal data, an important analytical task is unsupervised clustering, which aims to integrate all available modalities to uncover meaningful data patterns and provide insights. Various computational methods have been proposed to combine multiple modalities, extract shared and complementary information, and enhance clustering accuracy. For example, Rappoport and Shamir \cite{rappoport2018multi} provided a comprehensive review of multi-modal clustering methods, covering both omics-specific approaches and generic machine learning techniques. More recently, Leng et al. \cite{leng2022benchmark} benchmarked deep learning-based multi-modal integration methods, comparing their performance in clustering and classification tasks using multi-omics data. These studies highlight the importance of integrating multiple modalities for robust and  meaningful clustering results.

Integrative methods for multi-modal data are typically categorized into three main approaches: early, middle, and late integration \cite{flores2023missing}. Early integration concatenates all modalities into a single matrix, followed by a single-modality clustering method. A common strategy is to first apply principal component analysis (PCA) \cite{mackiewicz1993principal} or its variants for dimensionality reduction before performing clustering. 
Late integration first clusters each modality separately and then aggregates the clustering results. For example, the PINS method \cite{nguyen2017novel} constructs a similarity matrix by integrating connectivity information from each modality before refining the final clustering results. 

In contrast, middle integration represents a broad class of methods that jointly model all modalities within a unified framework. These approaches fall into two major categories: similarity-based integration and joint dimensionality reduction. Example similarity-based methods include SNF \cite{wang2014similarity}, which constructs sample-sample networks for each modality and iteratively fuses them using message passing, followed by spectral clustering. The NEMO method \cite{rappoport2019nemo} simplifies earlier similarity-based approaches by avoiding iterative optimization, and it enables application to datasets with missing modalities without requiring prior imputation.

Among joint dimensionality reduction methods, nonnegative matrix factorization (NMF) \cite{brunet2004metagenes} and its extensions \cite{liu2013multi} are widely used for extracting latent factors across multi-modal datasets. 
Canonical Correlation Analysis (CCA) \cite{hotelling1992relations} and its variants are another prominent class of joint dimensionality reduction techniques \cite{guo2019canonical}. CCA works by linearly projecting two modalities into a lower-dimensional space such that the correlation between them is maximized. Over time, multiple extensions of CCA have been developed to address different challenges. For example, sparse CCA incorporates a sparsity constraint to identify the most informative features \cite{witten2009extensions}. The extension to sparse MCCA generalizes CCA to handle more than two modalities, improving its ability to integrate multi-source data \cite{witten2009extensions}. Tilted-CCA  decomposes paired multi-modal data into lower-dimensional embeddings, which can separately quantify shared and modality-specific information \cite{lin2023quantifying}.


Another challenge in multi-modal data integration is the prevalence of missing values in real-world datasets, arising due to system limitations, data corruption, subject dropout, or budget constraints \cite{flores2023missing, pedersen2017missing}.
In early integration (i.e., concatenating all features into a single input space before modeling), probabilistic PCA (PPCA) provides a flexible framework for handling missing values using the Expectation-Maximization (EM) algorithm \cite{tipping1999probabilistic, severson2017principal, yu2010probabilistic}. 
In contrast, in CCA-based middle integration, an iterative imputation method was proposed to impute missing values without constructing a probabilistic model \cite{van2012generalized}.
In the specific problem of multi-omics data integration, where each omics type corresponds to one modality, the MOFA method \cite{argelaguet2018multi} adopts a variational Bayesian framework to simultaneously learn low-dimensional projections and impute missing values. However, its assumption of feature-wise independence may limit its ability to capture inter-feature correlations.

To address the challenge of multi-modal data integration with full or partial observations, we propose Generalized Probabilistic Canonical Correlation Analysis (GPCCA), an extension of probabilistic CCA \cite{bach2005probabilistic} that handles missing data across two or more modalities. GPCCA makes three key contributions: (1) it learns integrated low-dimensional embeddings in a unified probabilistic framework, (2) it inherently imputes missing values within its parameter estimation process, and (3) it identifies informative features while accounting for correlations between features within the same modality. To enhance numerical stability and model generalizability, we incorporate ridge regularization into the GPCCA model. An EM algorithm is developed to estimate model parameters. As detailed in the Results section, we evaluate GPCCA's performance on three datasets: a 3-modality simulated dataset under diverse settings, a 4-modality handwritten numeral image dataset, and a 3-modality omics dataset spanning diverse cancer types.

\section{Methods} \label{sec:mtds}
\subsection{The GPCCA model} \label{subsec:model}
Our GPCCA model generalizes the probabilistic CCA model from two modalities to $R$ ($R \geq 2$) modalities , while accommodating the presence of missing data. Suppose there are $n$ subjects on which $R$ modalities of data are measured. These modalities typically correspond to data collected using different technologies or experimental procedures. For the $r$-th modality, we denote the numeric data matrix (after appropriate pre-processing and transformation when applicable) as $\mathbf{X}^{(r)} \in \mathbb{R}^{m_r \times n}$, where $m_r$ represents the number of features in modality $r$. In GPCCA, we assume that each data modality can be factorized as follows:
\begin{equation}
\underset{m_r \times n}{\mb{X}^{(r)}} = 
\underset{m_r \times d}{\mb{W}^{(r)}} \cdot \underset{d \times n}{\mb{Z}} + 
\underset{m_r \times n}{\mb{U}^{(r)}} + \underset{m_r \times n}{\mb{E}^{(r)}}
,\ r=1,2,\dots,R,
\end{equation}
where 
${\mathbf{Z}} \in \mathbb{R}^{d \times n}$ is the latent embedding matrix corresponding to $d\ (1 \leq d \leq \min\{m_r\}_{1 \leq r \leq R})$ latent factors, with each column following a multivariate standard Gaussian distribution: $\{\mathbf{Z}_{\cdot k}\}_{1 \leq k \leq n} \overset{\text{i.i.d.}}{\sim} \mathcal{N}_d(\mathbf{0}, \mathbf{I})$. Notably, this low-dimensional hidden space with $d$ dimensions is shared by the $R$ modalities. $\mathbf{W}^{(r)} \in \mathbb{R}^{m_r \times d}$ is the loading matrix that measures how strongly each feature relates to each factor. $\mathbf{U}^{(r)} \in \mathbb{R}^{m_r \times n}$ is the matrix of mean vectors for uncentered data, defined as $\mathbf{U}^{(r)} = [\bm{\mu}^{(r)}, \bm{\mu}^{(r)}, \dots, \bm{\mu}^{(r)}]$, where the mean vector $\bm{\mu}^{(r)} \in \mathbb{R}^{m_r}$ is identical across all columns. $\mathbf{E}^{(r)} \in \mathbb{R}^{m_r \times n}$ is the matrix of random error terms, with each column following a multivariate Gaussian distribution: $\{\mathbf{E}^{(r)}_{\cdot k}\}_{1 \leq k \leq n} \overset{\text{i.i.d.}}{\sim} \mathcal{N}_{m_r}(\mathbf{0}, \bm{\Psi}^{(r)})$.

\begin{figure}
    \centering
    \includegraphics[width=\textwidth]{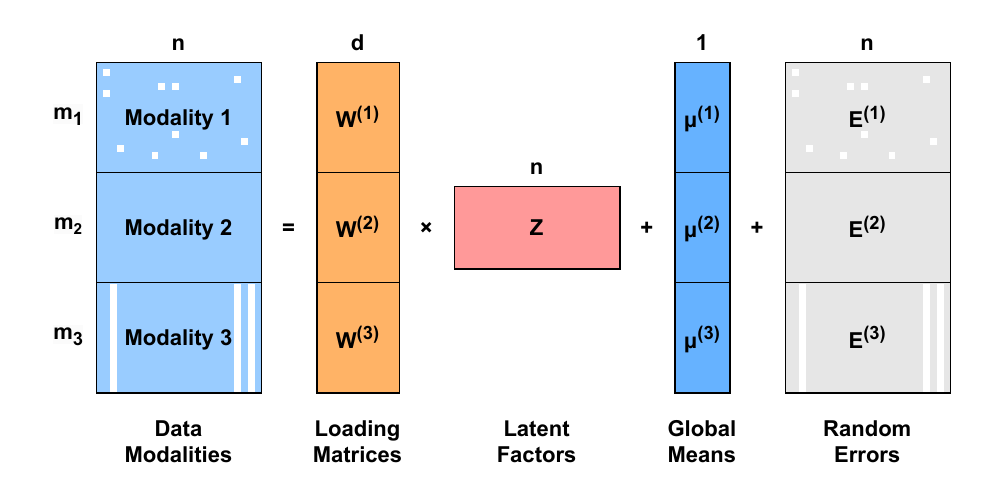}
    \caption{An illustration of the GPCCA model applied to a three-modality dataset. White boxes indicate missing data. In this example, Modality 2 is fully observed, while Modality 1 has randomly missing values, and Modality 3 exhibits modality-wise missingness.}
    \label{fig:GPCCA model}
\end{figure}


Combining all modalities together by stacking corresponding model components leads to the full model (Figure \ref{fig:GPCCA model}) as follows:
\begin{equation}
\begin{bmatrix} \mb{X}^{(1)} \\ \mb{X}^{(2)} \\ \vdots \\ \mb{X}^{(R)} \end{bmatrix} = 
\begin{bmatrix} \mb{W}^{(1)} \\ \mb{W}^{(2)} \\ \vdots \\ \mb{W}^{(R)} \end{bmatrix} \cdot
\mb{Z}+ 
\begin{bmatrix} \mb{U}^{(1)} \\ \mb{U}^{(2)} \\ \vdots \\ \mb{U}^{(R)} \end{bmatrix} + 
\begin{bmatrix} \mb{E}^{(1)} \\ \mb{E}^{(2)} \\ \vdots \\ \mb{E}^{(R)} \end{bmatrix}.
\end{equation}
For brevity, we present the full model using simplified notations:
\begin{equation}
\underset{m \times n}{\mb{X}} = 
\underset{m \times d}{\mb{W}} \cdot \underset{d \times n}{\mb{Z}} + 
\underset{m \times n}{\mb{U}} + \underset{m \times n}{\mb{E}},
\end{equation}
where $m = \sum_{r=1}^R m_r$ and $1 \le d \le \min\{m_r\}_{1 \le r \le R}$.
Next, we define $\mathbf{x}_k \coloneqq \mathbf{X}_{\cdot k}$, $\mathbf{z}_k \coloneqq \mathbf{Z}_{\cdot k}$, and $\bm{\epsilon}_k \coloneqq \mathbf{E}_{\cdot k}$ ($k=1,\dots,n$) for clarity. With these notations, our model assumptions are simplified as follows:
\begin{equation}
\{\mb{z}_k\}_{1 \le k \le n} \overset{\text{i.i.d.}}{\sim} \mathcal{N}_d(\mb{0}, \mb{I})\ \text{ and }\ 
\{\bm{\epsilon}_k\}_{1 \le k \le n} \overset{\text{i.i.d.}}{\sim} \mathcal{N}_m(\mb{0}, \bm{\Psi}), \text{ where } \bm{\Psi} = 
\begin{pmatrix}
  \bm{\Psi}^{(1)} & & & \\ 
  & \bm{\Psi}^{(2)} & & \\
  & & \ddots & \\
  & & & \bm{\Psi}^{(R)}
\end{pmatrix}.
\end{equation}
The error covariance matrix $\bm{\Psi}$ is block diagonal, allowing for potential correlations between error terms within the same modality, while assuming no correlations between error terms across modalities. This structure represents a trade-off between the overly simplistic diagonal covariance matrix and the fully parameterized covariance matrix, which may lead to overfitting and robustness issues.

\subsection{Estimation of the GPCCA model} \label{subsec:est}
We denote the complete set of parameters in the GPCCA model as $\bm{\Theta} = (\mathbf{W}, \bm{\mu}, \bm{\Psi})$. To accommodate incomplete data with missing values, we develop an EM algorithm \cite{dempster1977maximum} to estimate the model parameters. The EM algorithm relies on the assumption of missing at random (MAR), meaning that missingness depends only on the observed data and not on the missing values themselves. This assumption allows us to estimate model parameters without explicitly modeling the missing data mechanism.
We introduce the key steps of the algorithm below, and provide a more detailed derivation in Supplementary Methods 1.1. To distinguish between observed and missing values, we introduce an indicator matrix $\mb{O}\in\mathbb{R}^{m\times n}$, where
\begin{equation}
  \mb{O}_{ik} = \begin{cases}
    1, & \text{if $\mb{X}_{ik}$ is observed;} \\
    0, & \text{if $\mb{X}_{ik}$ is missing.}
  \end{cases}
\end{equation}
Starting with an initial set of parameters $\bm{\Theta}^{(0)} = (\mathbf{W}^{(0)}, \bm{\mu}^{(0)}, \bm{\Psi}^{(0)})$ (see Supplementary Methods 1.1 for parameter initialization), we denote the parameters obtained in the $t$-th iteration as $\bm{\Theta}^{(t)} = (\mathbf{W}^{(t)}, \bm{\mu}^{(t)}, \bm{\Psi}^{(t)})$. At the $(t+1)$-th iteration, we define the following key components for subject $k$ ($k = 1, 2, \dots, n$):
\begin{align}
\left\{
\begin{aligned}
\text{Partial (observed) data:} & \quad \Tilde{\mb{x}}_k \coloneqq [\mb{X}_{ik}]_{\{i : \mb{O}_{ik} = 1\}} \in \mathbb{R}^{m_{(k)}}
\\
\text{Partial loadings:} & \quad \Tilde{\mb{W}}_k^{(t)} \coloneqq [{\mb{W}_{i \cdot}^{(t)}}]_{\{i : \mb{O}_{ik} = 1\}} \in \mathbb{R}^{m_{(k)}\times d}
\\
\text{Partial means:} & \quad \Tilde{\bm{\mu}}_k^{(t)} \coloneqq [\bm{\mu}_i^{(t)}]_{\{i : \mb{O}_{ik} = 1\}}\in 
\mathbb{R}^{m_{(k)}} 
\\
\text{Partial error covariance:} & \quad \Tilde{\bm{\Psi}}_k^{(t)} \coloneqq \{\bm{\Psi}_{ij}^{(t)}\}_{\{(i,j) : \mb{O}_{ik} = 1, \mb{O}_{jk} = 1\}} 
\in 
\mathbb{R}^{m_{(k)}\times m_{(k)}} 
\\
\text{Woodbury identity matrix:} & \quad \Tilde{\mb{M}}_k^{(t)} \coloneqq (\mb{I} + {\Tilde{\mb{W}}_k^{(t)\text{T}}}(\Tilde{\bm{\Psi}}_k^{(t)})^{-1}\Tilde{\mb{W}}_k^{(t)})^{-1} \in 
\mathbb{R}^{d\times d} 
\end{aligned}
\right.{,}
\end{align}
where $m_{(k)} = |\{i : \mb{O}_{ik} = 1\}|$.
Then, in the E-step, we compute the following conditional expectations ($k=1,\dots, n, i=1,\dots,m, j=1,\dots,m,$):
\begin{align}\label{eq:Estep}
\begin{split}
\E(\mb{z}_k|\Tilde{\mb{x}}_k;\bm{\Theta}^{(t)}) &= \Tilde{\mb{M}}_k^{(t)}{\Tilde{\mb{W}}_k^{(t)\text{T}}}{\Tilde{\bm{\Psi}}_k^{(t)}}^{-1}(\Tilde{\mb{x}}_k-\Tilde{\bm{\mu}}_k^{(t)});
\\
\E(\mb{z}_k\mb{z}_k^{\text{T}}|\Tilde{\mb{x}}_k;\bm{\Theta}^{(t)}) &= \Tilde{\mb{M}}_k^{(t)} + \E(\mb{z}_k|\Tilde{\mb{x}}_k;\bm{\Theta}^{(t)})\E(\mb{z}_k|\Tilde{\mb{x}}_k;\bm{\Theta}^{(t)})^{\text{T}};
\\
\E(\mb{X}_{ik}|\Tilde{\mb{x}}_k;\bm{\Theta}^{(t)}) &= \begin{cases}
    \mb{X}_{ik}, & \text{if $\mb{O}_{ik}=1$}, \\
    \mb{W}_{i \cdot}^{(t)}\E(\mb{z}_k|\Tilde{\mb{x}}_k;\bm{\Theta}^{(t)}) + \bm{\mu}_i^{(t)}, & \text{if $\mb{O}_{ik}=0$};
  \end{cases}
\\
\E(\mb{X}_{ik}\mb{z}_k^{\text{T}}|\Tilde{\mb{x}}_k;\bm{\Theta}^{(t)}) &= \begin{cases}
    \mb{X}_{ik}\E(\mb{z}_k|\Tilde{\mb{x}}_k;\bm{\Theta}^{(t)})^{\text{T}}, & \text{if $\mb{O}_{ik}=1$}, \\
    \mb{W}_{i \cdot}^{(t)}\Tilde{\mb{M}}_k^{(t)} + \E(\mb{X}_{ik}|\Tilde{\mb{x}}_k;\bm{\Theta}^{(t)})\E(\mb{z}_k|\Tilde{\mb{x}}_k;\bm{\Theta}^{(t)})^{\text{T}}, & \text{if $\mb{O}_{ik}=0$};
  \end{cases}
\\  \E(\mb{X}_{ik}\mb{X}_{jk}|\Tilde{\mb{x}}_k;\bm{\Theta}^{(t)}) &= \begin{cases}
    \mb{X}_{ik}\mb{X}_{jk}, & \text{if $\mb{O}_{ik}=\mb{O}_{jk}=1$}, \\
    \E(\mb{X}_{ik}|\Tilde{\mb{x}}_k;\bm{\Theta}^{(t)})\mb{X}_{jk}, & \text{if $\mb{O}_{ik}=0, \mb{O}_{jk}=1$}, \\
    \mb{X}_{ik}\E(\mb{X}_{jk}|\Tilde{\mb{x}}_k;\bm{\Theta}^{(t)}), & \text{if $\mb{O}_{ik}=1, \mb{O}_{jk}=0$}, \\
    \mb{W}_{i \cdot}^{(t)}\Tilde{\mb{M}}_k^{(t)}{\mb{W}_{j \cdot}^{(t)\text{T}}} + \bm{\Psi}_{ij}^{(t)} 
    + \E(\mb{X}_{ik}|\Tilde{\mb{x}}_k;\bm{\Theta}^{(t)})\E(\mb{X}_{jk}|\Tilde{\mb{x}}_k;\bm{\Theta}^{(t)}),
    & \text{if $\mb{O}_{ik}=\mb{O}_{jk}=0$}.
  \end{cases}
\end{split}
\end{align}
In the M-step, we update parameters by the following formulas:
\begin{align}\label{eq:Mstep}
\begin{split}
\bm{\mu}^{(t+1)} &= \dfrac{1}{n}\sum\limits_{k=1}^n(\E(\mb{x}_k|\Tilde{\mb{x}}_k;\bm{\Theta}^{(t)}) - \mb{W}^{(t)}\E(\mb{z}_k|\Tilde{\mb{x}}_k;\bm{\Theta}^{(t)})),
\\
\mb{W}^{(t+1)} &= \left[\sum\limits_{k=1}^n(\E(\mb{x}_k\mb{z}_k^{\text{T}}|\Tilde{\mb{x}}_k;\bm{\Theta}^{(t)}) - \bm{\mu}^{(t)}(\E(\mb{z}_k))^{\text{T}}|\Tilde{\mb{x}}_k;\bm{\Theta}^{(t)})\right]\left[\sum\limits_{k=1}^n\E(\mb{z}_k\mb{z}_k^{\text{T}}|\Tilde{\mb{x}}_k;\bm{\Theta}^{(t)})\right]^{-1},
\\
\bm{\Psi}^{(t+1)} &= \dfrac{1}{n}\sum\limits_{k=1}^n \text{Bdiag}_{\bm{\Psi}}[{\mb{G}_k^{(t)\text{T}}}],
\end{split}
\end{align}
where
$
\mb{G}_k^{(t)} = \E(\mb{x}_k\mb{x}_k^{\text{T}}|\Tilde{\mb{x}}_k;\bm{\Theta}^{(t)}) + \bm{\mu}^{(t)}{\bm{\mu}^{(t)\text{T}}} + \mb{W}^{(t)}\E(\mb{z}_k\mb{z}_k^{\text{T}}|\Tilde{\mb{x}}_k;\bm{\Theta}^{(t)}){\mb{W}^{(t)\text{T}}} 
- 2\E(\mb{x}_k|\Tilde{\mb{x}}_k;\bm{\Theta}^{(t)}){\bm{\mu}^{(t)\text{T}}} + 2\mb{W}^{(t)}\E(\mb{z}_k|\Tilde{\mb{x}}_k;\bm{\Theta}^{(t)}){\bm{\mu}^{(t)\text{T}}} - 2\mb{W}^{(t)}\E(\mb{z}_k\mb{x}_k^{\text{T}}|\Tilde{\mb{x}}_k;\bm{\Theta}^{(t)})
$.
$\text{Bdiag}_{\bm{\Psi}}(\cdot)$ denotes a function that retains only the elements of its input matrix corresponding to the block diagonal structure of $\bm{\Psi}$, setting all off-block-diagonal entries to zero.
The E-step and M-step are performed iteratively until convergence. It is worth noting that the Woodbury identity matrix \cite{woodbury1950inverting} and the blockwise matrix inversion by modality are both introduced to improve the computational efficiency of the EM algorithm. In the case of complete data without any missing values, a simplified version of the EM algorithm is presented in Supplementary Methods 1.1. Both algorithms are offered in our GPCCA R package.

\subsection{Ridge regularization in the GPCCA model} \label{subsec:regularization}
As the dimensionality of features increases, estimating the error covariance matrix $\bm{\Psi}$ becomes increasingly unstable due to the high variance and sensitivity of standard covariance estimators in high-dimensional settings. This instability arises because the number of parameters to estimate in $\bm{\Psi}$ grows quadratically with the number of features, leading to overfitting and unreliable estimates, especially when the sample size is relatively small compared to the feature dimensionality. 
In many real-world applications, it is reasonable to assume that each feature is directly associated with only a limited subset of other features within each modality, implying a sparse or block-diagonal-like structure in the covariance matrix. 
Therefore, to address these challenges, we introduce ridge regularization on the error covariance matrix $\bm{\Psi}$. The ridge regularization we consider involves adding a penalty term proportional to the identity matrix, effectively shrinking the off-diagonal entries and stabilizing the estimation of of the error correlation matrix \cite{warton2008penalized}. By incorporating this regularization, we achieve a more robust and stable estimation of the error covariance matrix, improving the reliability of the model in both high-dimensional and limited-sample settings.

Suppose we use $\mb{R}$  to denote the error correlation matrix, then the error covariance matrix $\bm{\Psi}$ can be factorized as $\bm{\Psi} = \bm{\Psi}_d^{\frac{1}{2}}\mb{R}\bm{\Psi}_d^{\frac{1}{2}}$, where $\bm{\Psi}_d$ is a diagonal matrix with positive values. 
As derived in Supplementary Methods 1.1, the expected complete log-likelihood can be represented in the form below:
\begin{equation}
\E[\ell_{\text{c}}(\bm{\Theta})] \propto \mathcal{Q}(\bm{\Theta}) = -\dfrac{n}{2}\ln|\bm{\Psi}| - \dfrac{1}{2}\sum\limits_{k=1}^n\text{tr}(\bm{\Psi}^{-1}\mb{G}_k) - \dfrac{1}{2}\sum\limits_{k=1}^n\text{tr}(\E(\mb{z}_k\mb{z}_k^{\text{T}})).
\end{equation}
The ridge estimator of the error correlation matrix $\mb{R}$ is the maximum penalized likelihood estimator (MPLE) with a penalty term proportional to $-\text{tr}(\mb{R}^{-1})$ \cite{warton2008penalized}. Therefore, to obtain the ridge estimator, we modify $\mathcal{Q}(\bm{\Theta})$ by adding a penalty term as follows:
\begin{align}
\mathcal{Q}_{c}(\bm{R}) &= \mathcal{Q}(\bm{\Theta}) - \dfrac{c}{2}\text{tr}(\mb{R}^{-1})
\propto -\dfrac{n}{2}\ln|\bm{\Psi}| - \dfrac{1}{2}\sum\limits_{k=1}^n\text{tr}(\bm{\Psi}^{-1}\mb{G}_k) - \dfrac{c}{2}\text{tr}(\mb{R}^{-1}),
\end{align}
where $c\in(0, n)$ and terms not dependent on $\bm{R}$ are ignored. Then, based on the derivations in Supplementary Methods 1.2, we can show that the MPLE of the error correlation matrix is
\begin{equation}
\hat{\mb{R}}_{\text{ridge}} = \lambda\hat{\mb{R}} + (1-\lambda)\mb{I},
\end{equation}
where $\lambda = 1-\dfrac{c}{n}\in(0,1)$. Here, $\hat{\mb{R}} = \hat{\bm{\Psi}}_d^{-\frac{1}{2}}\hat{\bm{\Psi}}\hat{\bm{\Psi}}_d^{-\frac{1}{2}}$ denotes the original estimator of error correlation matrix without introducing the regularization term, and $\hat{\bm{\Psi}} = \dfrac{1}{n}\sum\limits_{k=1}^n \text{Bdiag}_{\bm{\Psi}}[\mb{G}_k^{\text{T}}]$ is the original MLE of $\bm{\Psi}$.
The corresponding ridge estimator of the error covariance matrix is
\begin{equation}
\hat{\bm{\Psi}}_{\text{ridge}} = \hat{\bm{\Psi}} + (\dfrac{1}{\lambda} - 1)\hat{\bm{\Psi}}_d.
\end{equation}
Because of the above relationships between the ridge estimators and the original estimators, this ridge regularization can be easily incorporated into the EM algorithm by updating the M-step in formula (\ref{eq:Mstep}).

The ridge estimator, $\hat{\mb{R}}_{\text{ridge}}$, effectively balances the bias-variance trade-off inherent in correlation matrix estimation. It acts as a weighted average of the sample correlation $\hat{\mb{R}}$ and the identity $\mb{I}$. In the shrinkage term, the choice of $\lambda$ (the regularization parameter) plays an important role in determining this balance.
The optimal value of the ridge regularization parameter can vary across datasets due to differences in dimensionality, correlation structures, and noise levels. When applying GPCCA to a new multi-modal dataset, multiple trials with varying $\lambda$ values can be performed in exploratory analysis to identify the most suitable parameter. 
Based on comprehensive simulation studies conducted on synthetic datasets, we observed that $\lambda = \frac{1}{2}$ and $\lambda=\frac{2}{3}$ consistently achieves good performance across diverse scenarios, suggesting them as robust default choices for the ridge regularization parameter.

\subsection{Selection of latent factor number} \label{subsec:hyperparam}

We introduce the method below to guide the selection of $d$, the number of latent factors. Consider a collection of candidate values $\{d_1, d_2, \dots, d_K\}$, ordered such that $d_i < d_j$ for all $1 \leq i < j \leq K$. For each candidate value \(d_k\), the EM algorithm for GPCCA is initialized with \(B\) different starting points, and consequently, $B$ sets of results are estimated for GPCCA given $d_k\ (k=1,\dots,K)$.
To evaluate the robustness of the model fitting with each candidate dimension, we perform Louvain clustering \cite{blondel2008fast} of subjects on the learned low-dimensional embeddings. The clustering results are aggregated into a matrix \(\mb{L}_k\in\mathbb{R}^{n \times B} \), where each column corresponds to clustering results from one initialization. Then, we compute a consensus matrix \(\mb{C}_k\in\mathbb{R}^{n\times n}\), where the \((i, j)\)-th entry reflects the proportion of clustering results in which subjects \(i\) and \(j\) are assigned to the same cluster.
To assess the consistency of the \(B\) sets of clusters resulting from the same latent dimension, we introduce the consensus score below, which provides a quantitative measure of the agreement between the multiple clustering results, offering insights into the reliability of the learned low-dimensional embeddings. The consensus score is formally defined as follows:
\begin{equation}
\mathcal{H}_k = \sum\limits_{i<j}\mb{C}_{k,ij}\log_2(\mb{C}_{k,ij}),
\end{equation}
which has an upper bound of $0$ when the $B$ sets of clustering results are exactly the same.
The final number of latent factors $d_{k^*}$ is selected as the candidate value which corresponds to the largest consensus score. Then, the last step is to find the best initialization $b\in\{1,2,\dots,B\}$. For each initialization, we can compute the binary connectivity matrix $\mb{C}_{k^*}^{(b)}$, where the $(i,j)$-th entry is 1 if subjects $i$ and $j$ belong to the same cluster. We compare this connectivity matrix with the consensus matrix based on the root mean squared error (RMSE):
\begin{equation}
\text{RMSE}_b = \sqrt{\frac{2}{n(n-1)}\sum\limits_{i<j}\left(\mb{C}^{[b]}_{k^*,ij} - \mb{C}_{k^*,ij}\right)^2}.
\end{equation}
Finally, we select the initialization that leads to the smallest RMSE.

\subsection{Design of simulation study} \label{subsec:sim designs}
In order to evaluate the performance of GPCCA in a realistic setting but with ground truth information, we generate synthetic multi-modal data reflecting different scenarios. In this simulation, we consider a three-modality design, where
the original dimensions of the three modalities are $m_1 = 60$, $m_2 = 120$, and $m_3 = 180$. In addition, we assume that there are six real clusters, where each cluster has a sample size of $100$. Each modality provides partial information of the clustering relationships. Within each modality, we consider both informative features that help distinguish between different clusters and noisy features that do not contribute to cluster separation. In all simulation settings, we set the signal to noise ratio (i.e., ratio between informative and noisy features within modality) to $1:4$. Below we introduce how the synthetic data is generated under four different cases.


\begin{table}
\centering
\caption{Data distributions of informative features in the simulation study.}
\begin{tabular}{ |p{4cm}|c|c|c| }
 \hline
 \multicolumn{1}{|c}{} & \multicolumn{3}{|c|}{Data Modality} \\
 \hline
 Observations $(n = 600)$ & $\mb{X}^{(1)}$: $m_1 = 60$ & $\mb{X}^{(2)}$: $m_2 = 120$ & $\mb{X}^{(3)}$: $m_3 = 180$ \\
 \hline
 Cluster 1 ($n_1 = 100$) & $f_u^{(1)}$ & $f_u^{(2)}$ & $f_u^{(3)}$ \\
 Cluster 2 ($n_2 = 100$) & $f_v^{(1)}$ & $f_v^{(2)}$ & $f_u^{(3)}$ \\
 Cluster 3 ($n_3 = 100$) & $f_u^{(1)}$ & $f_v^{(2)}$ & $f_u^{(3)}$ \\
 Cluster 4 ($n_4 = 100$) & $f_v^{(1)}$ & $f_u^{(2)}$ & $f_v^{(3)}$ \\
 Cluster 5 ($n_5 = 100$) & $f_v^{(1)}$ & $f_u^{(2)}$ & $f_u^{(3)}$ \\
 Cluster 6 ($n_6 = 100$) & $f_u^{(1)}$ & $f_v^{(2)}$ & $f_v^{(3)}$ \\
 \hline
\end{tabular}
\label{tab:compact design}
\end{table}



\textbf{Case A: normal data (MCAR)}. In Case A, we assume that data follows a normal distribution with missing completely at random (MCAR) mechanism. Below we describe different procedures to generate data for informative features and noisy features. For the informative features, within the $r$-th ($r=1,2,3$) modality, we pick three clusters where observations follow a multivariate normal distribution $f^{(r)}_u = \mathcal{N}(\bm{\mu}^{(r)}_u, \bm{\Sigma}^{(r)})$, while in the other three clusters, observations follow a different multivariate normal distribution $f^{(r)}_v = \mathcal{N}(\bm{\mu}^{(r)}_v, \bm{\Sigma}^{(r)})$ (Table \ref{tab:compact design}). Elements in the mean vectors $\bm{\mu}^{(r)}_u$ and $\bm{\mu}^{(r)}_v$ are independently drawn from uniform distributions $U[1, 2]$ and $U[-2, -1]$, respectively. The covariance matrix is computed as $\bm{\Sigma}^{(r)} = \mb{D}^{(r)}\bm{\Sigma}_{\text{AR}(1)}(\rho)\mb{D}^{(r)}$,  where $\mb{D}^{(r)} = \text{Diag}[\bm{\sigma}^{(r)}]$, and $\bm{\sigma}^{(r)}$ is drawn from $4\text{Beta}(1,1)$. $\bm{\Sigma}_{\text{AR}(1)}(\rho)$ is a correlation matrix with an order-1 autoregressive structure, where the $(i,j)$-th entry is $\rho^{|i-j|}$. In our study, different values of $\rho \in \{0.3, 0.5, 0.7, 0.8, 0.9\}$ are used to generate data with varying correlation strength.
In contrast, the generation of noisy features is independent of the cluster membership. Each noisy feature is independently sampled from a standard normal distribution. In addition, to evaluate the robustness of the method under varying levels of missingness, we randomly introduce missing values into the generated data. We consider three scenarios with different proportions of missing data: 0\% (complete data), 20\%, and 40\%.

\textbf{Case B: heavy-tailed data (MCAR).} 
In Case B, we consider heavy-tailed distribution where the assumption of normal distribution in the GPCCA model is violated. The simulation procedure is similar to that of Case A, except that the normal distribution is replaced with the $t$ distribution.
For the informative features, within the $r$-th ($r=1,2,3$) modality, we pick three clusters where observations follow a multivariate $t$ distribution with $3$ degress of freedom: $f^{(r)}_u = t_3(\bm{\mu}^{(r)}_u, \bm{\Sigma}^{(r)})$, while in the other three clusters, observations follow a different multivariate $t$ distribution $f^{(r)}_v = t_3(\bm{\mu}^{(r)}_v, \bm{\Sigma}^{(r)})$. The mean vectors and covariance matrices are generated as described in Case A.  As for the noisy features, each feature is independently sampled from a $t_3$ distribution. In addition, we also randomly introduce missing values into the generated data with different proportions of missing data: 0\% (complete data), 20\%, and 40\%.

\textbf{Case C: normal data (MNAR).} 
In Case C, we consider a scenario where the missingness is not at random (MNAR). First, we generate complete data for the three modalities as described in Case A. Then, we introduce modality-specific missingness based on the behavior of a hidden variable. The hidden variable $H_k$ determines whether the entire set of features within a modality is missing for a given sample $k\ (k=1,\dots,n)$. This design mimics real-world biological or biomedical scenarios where the availability of experimental results for a modality depends on latent factors, such as logistical constraints, sample preparation quality, or inherent biological characteristics of the subject. Specifically, for sample $k$, we sample $H_k$ from a standard normal distribution. If $H_k \ge 0$, the probability that this sample missing an entire modality is $p$; otherwise, if $H_k < 0$, the probability that this sample missing an entire modality is $2p$. We then introduce missing values into the data based on these probabilities. We generate two sets of data with $p = 0.1$ and $p = 0.2$, respectively. It is important to note that while $H_k$ values are generated for simulation purposes, they are considered hidden variables during data analysis and remain unobservable to the models. For example, Supplementary Figures S1 and S2 visualize the missing patterns in synthetic data with $p = 0.1$ and $p = 0.2$, respectively.

\textbf{Case D: correlated modalities (MCAR).}
In Case D, we consider data with dependence between modalities, which violates the assumption in GPCCA that there is no correlation between features across modalities. To achieve this goal, we first simulate the block diagonal error covariance matrix (for informative features) as described in Case A. Then, $1/6$ of the within-modality correlation entries are randomly selected to be swapped with between-modality correlation entries which are zeros. The other steps in the data generation approach remain the same as Case A. For example, Supplementary Figure S3 visualizes the final covariance matrices generated in Case A and Case D with $\rho = 0.9$.

\subsection{Alternative methods and comparative analysis} \label{subsec:alt mtds}
In the simulation study and real data applications, we consider GPCCA alongside four alternative methods for comparison, including PPCA \cite{roweis1997algorithms}, MOFA \cite{argelaguet2018multi}, SNF \cite{wang2014similarity}, and NEMO \cite{rappoport2019nemo}. The above methods are directly applied to the complete data. For incomplete data with missing values, SNF and NEMO cannot be directly applied, and missing values are first imputed using the k-nearest neighbors (kNN) imputation with $5$ nearest neighbors. GPCCA, PPCA, and MOFA can handle the missing values without the need of prior imputation.

To obtain clustering results, for methods capable of learning low-dimensional projections, including GPCCA, PPCA, and MOFA, we apply the Louvain clustering algorithm \cite{blondel2008fast} to the projected data with a default resolution parameter of 0.8. GPCCA and MOFA are specifically designed to account for the multi-modal structure of the data, whereas PPCA is implemented either on individual modalities or on a single dataset formed by concatenating multiple modalities. In contrast, similarity-based methods, including SNF and NEMO, compute similarity matrices for each modality, integrate these matrices, and then apply spectral clustering \cite{ng2001spectral} to the final similarity matrix to identify clusters. 
When ground-truth cluster labels are available, clustering performance is evaluated using the Adjusted Rand Index (ARI) \cite{yeung2001details}. ARI quantifies the agreement between the inferred cluster assignment and the true cluster assignment in the data. It ranges between $-1$ and $1$, with $1$ indicating a complete agreement between the inferred clusters and the ground truth. 

In the simulation study, the number of latent factors in GPCCA is selected from $\{2, 3, 4, 6, 8, 10\}$ using the approach introduced in the ``Selection of latent factor number'' section. PPCA is implemented with the number of principal components (PCs) set to $5$ for both single-modal and multi-modal data. MOFA is applied with its default parameter settings. The number of clusters in spectral clustering is set to the true cluster number ($6$) for SNF as an automatic selection method is not available. In contrast, NEMO determines the number of clusters automatically as part of its algorithmic procedure. 

In real data applications, the ridge regularization parameter is set to $\lambda = \frac{1}{2}$, and the number of latent factors in GPCCA is selected from $\{5, 10, 15, 20, 25, 30\}$.
For PPCA, the number of PCs is selected based on cross-validation (using the pcaMethods package \cite{stacklies2007pcamethods}) in the multi-view image data application. For the TCGA application, the elbow method is used to select the number of PCs (using the findPC package \cite{zhuang2022findpc}) as the cross-validation approach is too time-consuming. MOFA is applied with its default parameter settings. The number of clusters for SNF is set to the ground truth in the multi-view image data application.

\subsection{Real datasets} \label{subsec:RWD apps}

\textbf{Multi-view data of handwritten numerals.} 
The dataset consists of four feature sets of handwritten numerals ($0$ to $9$) extracted from a collection of Dutch utility maps \cite{multiple_features_72}. There are totally $2000$ observations, with $200$ observations in each class. Each feature set is obtained using a different feature extraction method for images, and we treat each of them as one modality. We use the following four modalities  in our analysis:
Modality 1 ($m_1 = 76$) contains Fourier coefficients of the character shapes;
Modality 2 ($m_2 = 216$) contains profile correlations;
Modality 3 ($m_3 = 64$) contains Karhunen-Loève coefficients;
Modality 4 ($m_4 = 47$) contains Zernike moments.
A more detailed introduction to features in these modalities is provided in Supplementary Methods 1.3.
In addition, there are no missing values in the original data.
To assess method performance in the presence of missing data, we generate four sets of additional data with missing values using the following procedure.

First, we introduce missing values completely at random at rates of 20\% and 40\%, resulting in two datasets: MCAR 20\% and MCAR 40\%.
Next, we introduce modality-specific missingness using a sequence of hidden variables,  $\{H_k\}_{1 \leq k \leq n} \overset{\text{i.i.d.}}{\sim} \mathcal{N}(0, 1)$, which determines whether an entire modality is missing for each sample \( k \) (\( k = 1, \dots, n \)). Specifically, if \( H_k \geq 0 \), the sample has a probability \( p \) of missing an entire modality. Otherwise, if \( H_k < 0 \), the probability increases to \( 2p \).
We then introduce missing values into the complete dataset according to these probabilities. By setting \( p = 0.25 \) and \( p = 0.50 \), we generate two additional datasets: MNAR 25\% and MNAR 50\%.

\textbf{Multi-omics data from the TCGA database.}
From The Cancer Genome Atlas (TCGA) \cite{cancer2008comprehensive} database, we  select 10 cancer types and downloaded multi-omics data of three modalities: gene expression levels (Illumina HiSeq), DNA methylation levels (Methylation450k), and miRNA expression levels (Illumina Hiseq). 
These cancer types were selected based on the following criteria. First, all three data modalities should be available for the cancer type. Second, the phenotype data (survival time) should be available for at least 100 tumor samples. Third, each data modality should contain at least 50 tumor samples.
As a result, the 10 selected cancer types include lung adenocarcinoma (LUAD), colon cancer (COAD), lung squamous cell carcinoma (LUSC), sarcoma (SARC), bladder cancer (BLCA), breast cancer (BRCA), head and neck cancer (HNSC), kidney clear cell carcinoma (KIRC), lower grade glioma (LGG), and liver cancer (LIHC). After data preprocessing (see Supplementary Methods 1.4), the sample sizes of each cancer type are summarized in Supplementary Table S1.

\section{Results} 

\subsection{Simulation}\label{sec:sim results}

To assess the performance of GPCCA in a controlled yet realistic setting with known ground-truth information, we generate four cases of synthetic multi-modal data characterized by distinct distributions and missing mechanisms (see Methods). The four cases are defined as follows: Case A involves normally distributed data with missing completely at random (MCAR); Case B considers heavy-tailed data with MCAR; Case C explores normally distributed data with missing not at random (MNAR); and Case D examines scenarios with correlated modalities under MCAR. Importantly, Cases B-D represent situations where the underlying assumptions of GPCCA are violated in various ways. These cases are designed to evaluate the robustness of GPCCA and the quality of its inferred data embeddings under such challenging conditions. In each case, we generate datasets comprising three modalities and evaluate the performance of GPCCA (regularizatrion parameter $\lambda = 2/3\ \text{or}\ 1/2$) alongside four alternative methods: PPCA, MOFA, SNF, and NEMO (see Methods for implementation). For dimensionality reduction approaches (GPCCA, PPCA, and MOFA), in order to evaluate the clustering performance, Louvain clustering is employed on their low-dimensional projections. To ensure a comprehensive assessment of method performance, we consider three levels of missingness ($0, 20\%, 40\%$) and five levels of feature correlation controlled by the parameter $\rho\in\{0.3, 0.5, 0.7, 0.8, 0.9\}$.

\begin{figure}
    \centering
    \includegraphics[width=\textwidth]{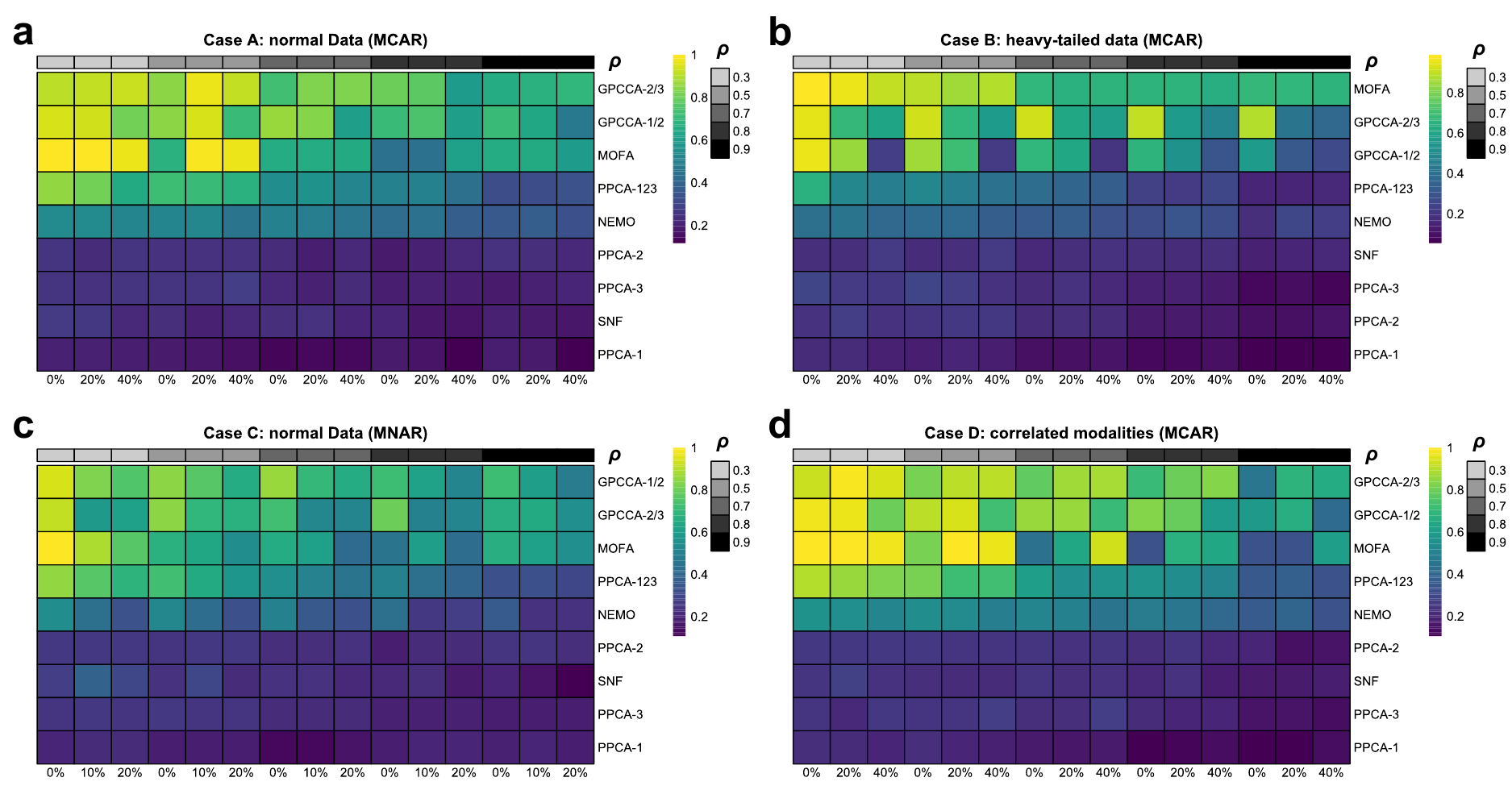}
    \caption{
    Comparison of clustering performance (ARI) in simulation study. \textbf{a.} Case A: normal data (MCAR). \textbf{b.} Case B: heavy-tailed data (MCAR). \textbf{c.} Case C: normal data (MNAR).
    \textbf{d.} Case D: correlated modalities (MCAR). GPCCA with different regularization parameters are denoted as GPCCA-2/3 and GPCCA-1/2. PPCA applied to concatenated modalities is denoted as PPCA-123. PPCA applied to individual modalities is denoted as PPCA-1, PPCA-2, and PPCA-3, respectively. Methods are ordered from high to low, based on their average ARI across all scenarios with different missing levels and correlation levels. In Cases A, B, and D, the horizontal labels on the bottom of the heatmap represent missing rates; in Case C, the horizontal labels represent the baseline probability ($p$) of modality-wise missingness (see Methods).
    }
    \label{fig:sim - selected ARIs}
\end{figure}

\begin{figure}
    \centering
    \includegraphics[width=\textwidth]{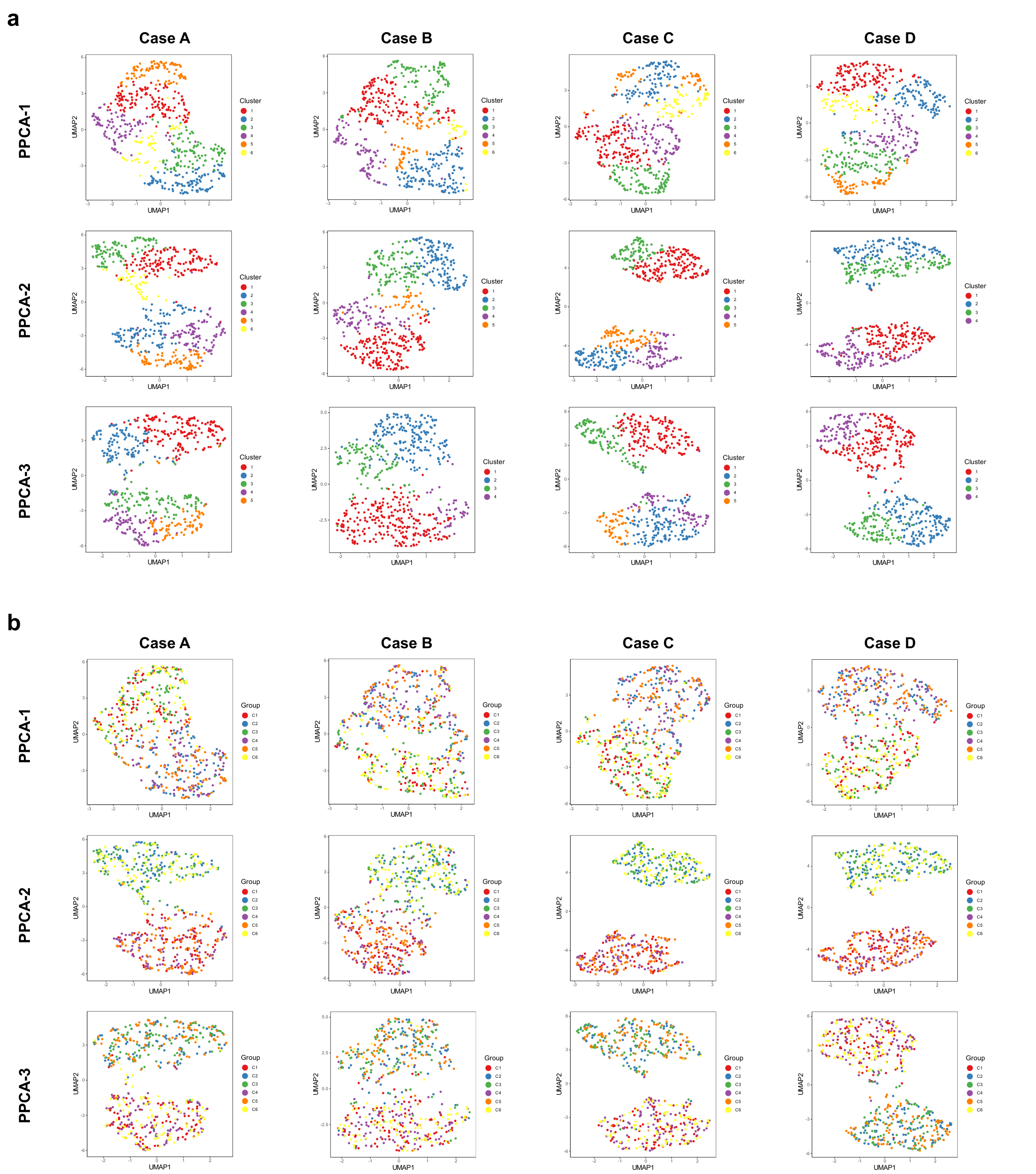}
    \caption{
    UMAP projections based on PPCA applied to every single modality. \textbf{a.} Samples are colored by the inferred clusters. \textbf{b.} Samples are colored by the true group labels. The demonstrated parameter settings are as follows: Cases A, B, and D ($20\%$ missing rate and $\rho = 0.7$); Case C (modality missingness with $p = 0.1$ and $\rho = 0.7$).
    }
    \label{fig:sim - UMAPs rho = 0.7 (PPCA only)}
\end{figure}

\begin{figure}
    \centering
    \includegraphics[width=\textwidth]{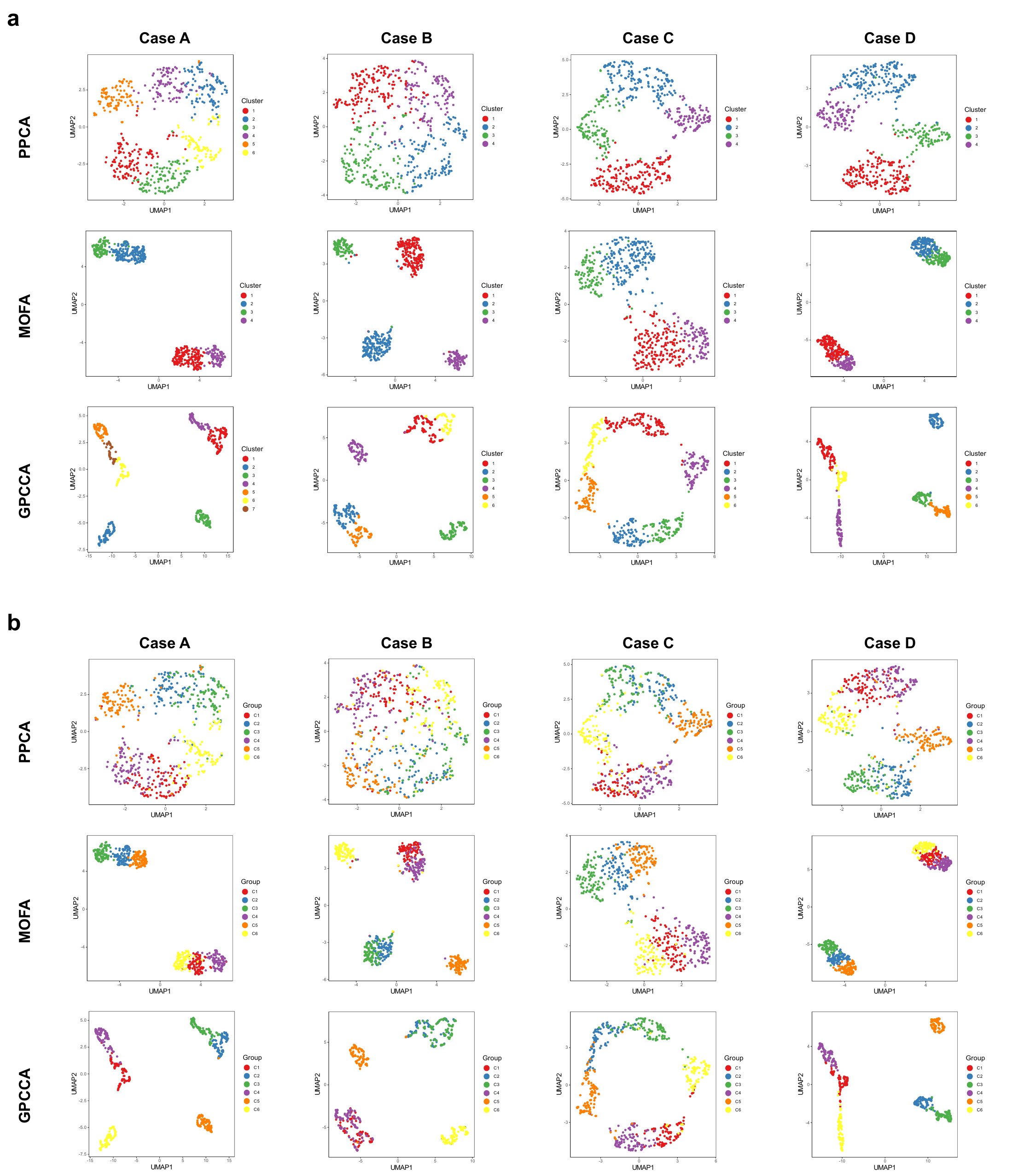}
    \caption{
    UMAP projections based on multi-modality analysis by PPCA, MOFA, and GPCCA. \textbf{a.} Samples are colored by the inferred clusters. \textbf{b.} Samples are colored by the true group labels. The demonstrated parameter settings are as follows: Cases A, B, and D  ($20\%$ missing rate and $\rho = 0.7$); Case C (modality missingness with $p = 0.1$ and $\rho = 0.7$). For GPCCA results, the model with better ARI score is used for visualization: $\lambda = 2/3$ for Case A, B and D  and $\lambda = 1/2$ for Case C.
    }
    \label{fig:sim - UMAPs rho = 0.7}
\end{figure}

In Case A, we analyze normal data with an MCAR mechanism. The clustering performance, as assessed by ARI scores, is summarized in Figure \ref{fig:sim - selected ARIs}a. To facilitate a direct comparison between single-modality and multi-modality analyses, PPCA is applied both to data from individual modalities and to concatenated multi-modality data. Since each modality provides only partial information about cluster membership, PPCA applied to the concatenated data consistently outperforms its application to individual modalities. For example, when missing rate is $20\%$ and $\rho=0.7$, PPCA over concatenated modalities consistently provides better separation of identified clusters and higher alignment with the true group labels, than PPCA over every single modality (Figures \ref{fig:sim - UMAPs rho = 0.7 (PPCA only)}-\ref{fig:sim - UMAPs rho = 0.7}). Overall, multi-modal methods outperform single-modality methods in clustering tasks, with GPCCA ($\lambda = 2/3$) achieving the best performance, followed by GPCCA ($\lambda = 1/2$), MOFA, and PPCA-123 (PPCA applied to concatenated data). For most methods, an increase in missing data leads to greater challenges in clustering. However, GPCCA demonstrates relatively stable performance across both complete and partial data. Additionally, as within-modality correlations (represented by $\rho$) increase, clustering becomes more difficult for all methods. For low correlation ($\rho = 0.3$), MOFA yields the best ARI score, as it does not account for correlations in the error covariance matrix. In contrast, for intermediate to high correlations ($\rho > 0.5$), GPCCA with ridge regularization exhibits superior performance. In the example with $20\%$ missing rate and $\rho=0.7$ (Figure \ref{fig:sim - UMAPs rho = 0.7}), GPCCA ($\lambda = 2/3$) achieves an ARI of $0.829$, identifying clusters that align well with the true clusters. In contrast, MOFA achieves an ARI of $0.664$, unable to distinguish between clusters C2 and C5, or C1 and C6. UMAP visualizations given other correlation settings are provided in Supplementary Figures S4-S7.

In Case B, we consider heavy-tailed data where the multivariate normal distribution used in Case A is replaced with a multivariate $t$ distribution. Heavy-tailed distributions are characterized by an increased likelihood of extreme values, which can lead to more ambiguous separation between clusters. As a result, we observe a general trend of decreasing ARI scores in Case B (Figure \ref{fig:sim - selected ARIs}b) compared with Case A, as the clustering becomes more challenging due to the influence of outliers and extreme values.
In this case, MOFA achieves the best average ARI, followed by GPCCA ($\lambda = 2/3$), GPCCA ($\lambda = 1/2$), and PPCA-123. Notably, similar to Case A, the advantage of MOFA is most pronounced in scenarios with low within-modality correlation. However, as the within-modality correlation increases, MOFA and GPCCA ($\lambda = 2/3$) demonstrate similar clustering performance, with GPCCA showing significantly better accuracy when the data is complete.

In Case C, we analyze data with MNAR patterns, specifically focusing on modality-wise missingness. In this scenario, certain subjects lack data for an entire modality, and the probability of missingness depends on an unobserved hidden variable. This setup reflects real-world challenges where the availability of data for a modality is influenced by certain latent factors, such as logistical constraints or inherent biological characteristics. For instance, in biomedical research, the presence of specific omics data may depend on the subject's condition or resource availability.
Compared to the MCAR mechanism in Case A, inferring cluster membership becomes more challenging under the MNAR condition, as evidenced by the overall lower ARI scores (Figure \ref{fig:sim - selected ARIs}c). Despite this, GPCCA ($\lambda = 1/2$) demonstrates the best average ARI performance, followed by GPCCA ($\lambda = 2/3$), MOFA, and PPCA-123. For example, under conditions where $\rho = 0.7$ and the baseline modality missing probability is $p = 0.1$ (Figure \ref{fig:sim - UMAPs rho = 0.7}), GPCCA ($\lambda = 1/2$) achieves an ARI of $0.701$, successfully identifying six clusters. In contrast, MOFA (ARI $= 0.613$) and PPCA (ARI $= 0.504$) identify only four clusters.
This case highlights the robustness of GPCCA in clustering analysis under the MNAR condition, where its model assumption is violated.

In Case D, we analyze data exhibiting across-modality correlations (i.e., correlations between features in different modalities), which violates the assumption of a block diagonal covariance matrix in GPCCA. In this scenario, GPCCA ($\lambda = 2/3$) achieves the best average ARI performance, followed by GPCCA ($\lambda = 1/2$), MOFA, and PPCA-123 (Figure \ref{fig:sim - selected ARIs}d). Compared to Case A, both GPCCA and MOFA demonstrate stable clustering performance even in the presence of across-modality correlations.
In summary, the simulation study highlights GPCCA's accuracy and robustness in clustering observations from multi-modality data. Although the model is developed under the assumptions of normal data and an MAR mechanism, it exhibits strong potential for extension to more complex scenarios involving heavy-tailed data or an MNAR mechanism.

\subsection{Application to multi-view image data} \label{sec:RWD app 2 results}

\begin{figure}
    \centering
    \includegraphics[width=18cm]{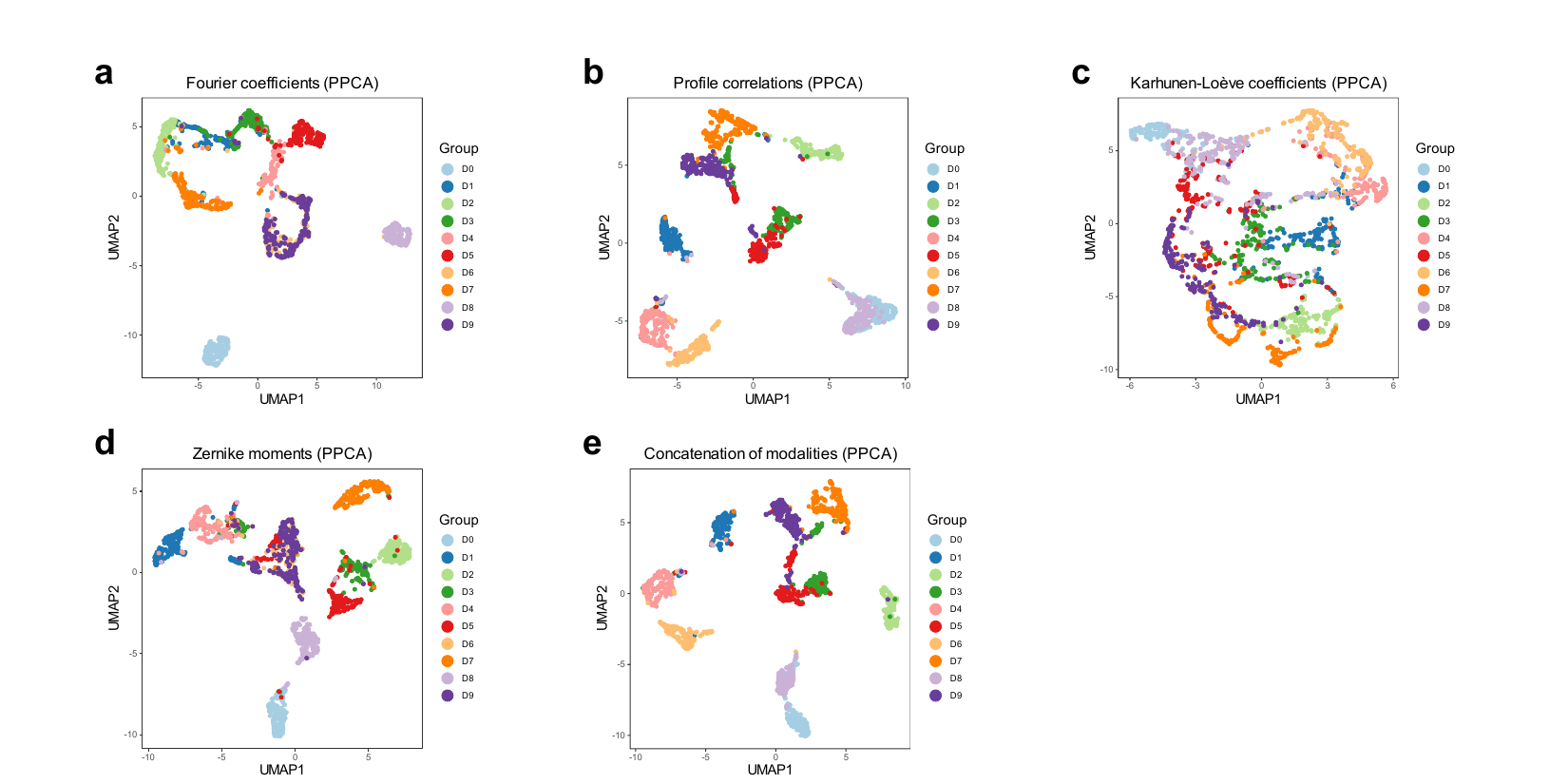}
    \caption{UMAP projections of PPCA results based on the four individual modalities and their concatenated data.
    \textbf{a.} Fourier coefficients.
    \textbf{b.} Profile correlations.
    \textbf{c.} Karhunen-Loève coefficients.
    \textbf{d.} Zernike moments.
    \textbf{e.} Concatenated data of the four modalities. Samples are colored by the true class labels.}
    \label{fig:mf10HD - PPCA by modality}
\end{figure}

In this section, we apply GPCCA to a multi-view dataset of handwritten numerals (see Methods) \cite{multiple_features_72}, which includes ground truth labels and four modalities corresponding to different numerical feature extraction methods applied to the images. The features within these modalities exhibit varying strengths in their ability to differentiate the digits (Supplementary Figures S8–S11). We hypothesize that jointly modeling features across modalities will enhance the model's capacity to distinguish between different classes.
%
%
%
To investigate this question, we first assess whether a single modality is sufficient to distinguish the 10 digits (0–9) by applying PPCA separately to each modality (Figure \ref{fig:mf10HD - PPCA by modality}a-d). The results highlight challenges in distinguishing certain digits depending on the modality. For instance, Fourier coefficients and Zernike moments struggle to differentiate between 6 and 9, while profile correlations fail to separate 0 from 8 and have difficulty distinguishing 3 from 5. Similarly, Karhunen-Loève coefficients do not generate well-separated clusters in the low-dimensional space for multiple digits. In contrast, when PPCA is applied to the concatenated data from all four modalities (Figure \ref{fig:mf10HD - PPCA by modality}e), the UMAP plot reveals improved separation between digits, suggesting that integrating multiple modalities enhances discriminability.

\begin{figure}
    \centering
    \includegraphics[width=\textwidth]{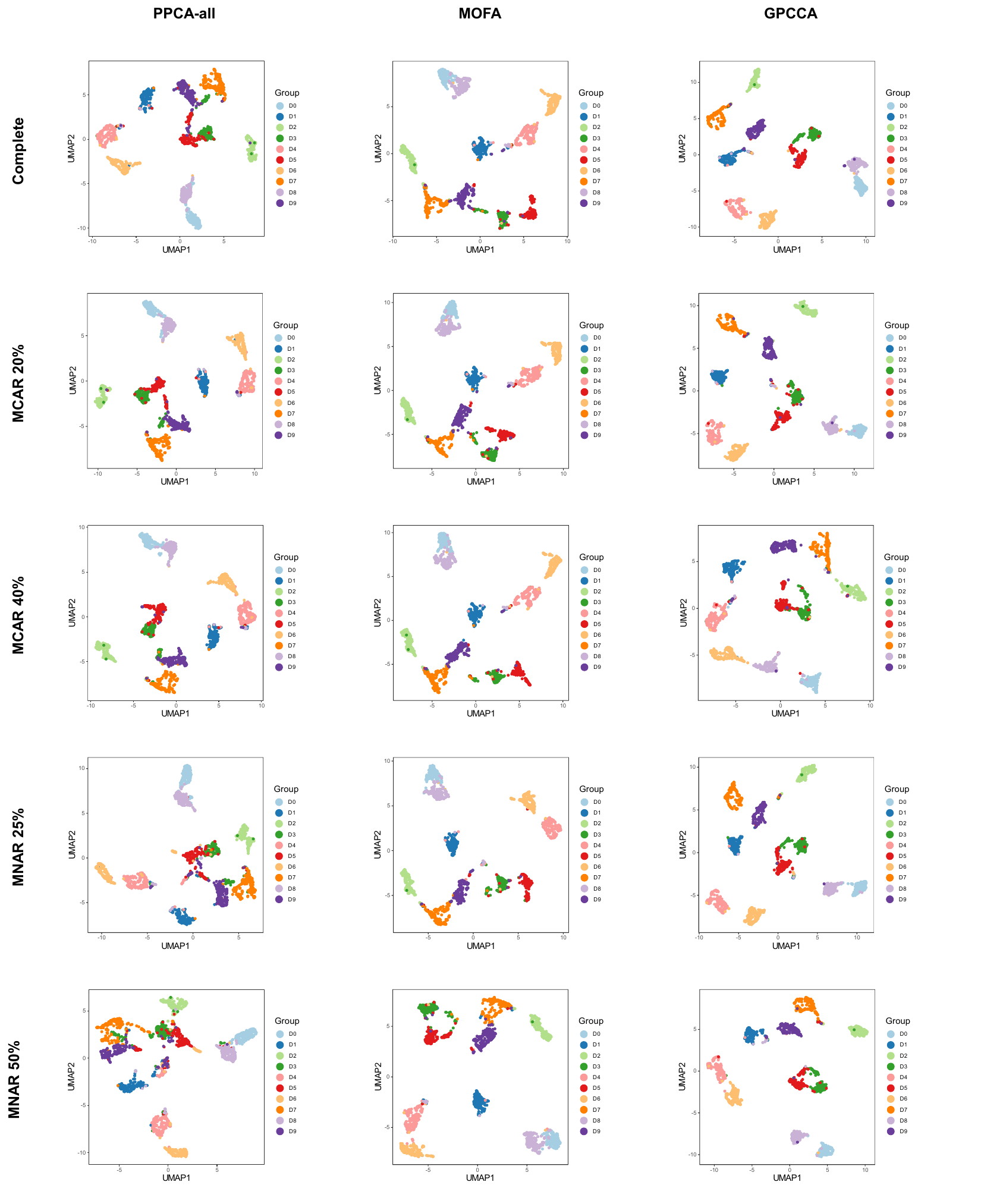}
    \caption{UMAP projections based on PPCA-all, MOFA, and GPCCA the multi-modality image data. The columns correspond to different methods, and the rows correspond to the complete data and four sets of partial data with missing values, respectively. Samples are colored by the true class labels.}
    \label{fig:mf10HD - results of studies}
\end{figure}

We next apply GPCCA to the multi-modality image data and compare its performance with alternative methods, including MOFA, SNF, NEMO, and PPCA. On the complete dataset, GPCCA achieves the highest ARI score for clustering results, followed by MOFA and NEMO (Table \ref{tab:mf10HD - ARIs}). Additionally, most multi-modality methods outperform single-modality methods in clustering accuracy. Compared to PPCA applied to individual modalities, GPCCA improves the distinction between digits 6 and 9 (Figure \ref{fig:mf10HD - results of studies}) and performs well in other challenging cases, such as separating 0 from 8 and 3 from 5.

\begin{table}[]
\centering
\caption{ARI scores of clustering results using different analytical methods. PPCA-all, GPCCA, MOFA, SNF, and NEMO are applied to the multi-modality data, whereas PPCA-fou, PPCA-fac, PPCA-kar, and PPCA-zer represent PPCA applied to Fourier coefficients, Profile correlations, Karhunen-Loève coefficients, and Zernike moments, respectively. For each dataset, the highest ARI value is highlighted in bold.
}
\begin{tabular}{ |p{3cm}|p{2cm}|p{2cm}|p{2cm}|p{2cm}|p{2cm}| }
 \hline
 Methods & Complete & MCAR 20\% & MCAR 40\% & MNAR 25\% & MNAR 50\% \\
 \hline
 PPCA-fou & 0.6284 & 0.6453 & 0.5584 & 0.5565 & 0.6839 \\
 PPCA-fac & 0.6914 & 0.6660 & 0.6692 & 0.6602 & 0.6462 \\
 PPCA-kar & 0.4120 & 0.5170 & 0.6284 & 0.1382 & 0.1525 \\
 PPCA-zer & 0.6499 & 0.6231 & 0.5411 & 0.6287 & 0.6344 \\
 PPCA-all & 0.7481 & 0.7507 & 0.7716 & 0.7095 & 0.6560 \\
 GPCCA & \textbf{0.8697} & \textbf{0.9072} & \textbf{0.8354} & \textbf{0.9090} & \textbf{0.9008} \\
 MOFA & 0.8408 & 0.8213 & 0.8344 & 0.8633 & 0.8391 \\
 SNF & 0.6114 & 0.4790 & 0.4745 & 0.5325 & 0.4972 \\
 NEMO & 0.7788 & 0.6940 & 0.6601 & 0.7737 & 0.7620 \\
 \hline
\end{tabular}
\label{tab:mf10HD - ARIs}
\end{table}

To further evaluate GPCCA’s robustness in the presence of missing data, we introduce missing values under different mechanisms and rates, generating four additional multi-modality image datasets: MCAR 20\%, MCAR 40\%, MNAR 25\%, and MNAR 50\% (see Methods). Applying GPCCA and alternative methods to these datasets shows that GPCCA consistently achieves the highest ARI, regardless of whether missing values are introduced at random (Table \ref{tab:mf10HD - ARIs} and Figure \ref{fig:mf10HD - results of studies}). Among the multi-modality methods, PPCA, GPCCA, MOFA, and NEMO exhibit stable performance on missing-data scenarios compared to their results on the complete data. In contrast, SNF suffers a significant drop in accuracy when applied to imputed data. These results highlight GPCCA’s ability to effectively integrate complementary information across multiple modalities and produce stable clustering outcomes.

\subsection{Application to multi-omics data} \label{sec:RWD app 1 results}

In this section, we analyze multi-omics data from The Cancer Genome Atlas (TCGA) database. We select 10 cancer types, each with three modalities: gene expression, DNA methylation, and miRNA expression (see Methods). For dimensionality reduction, we apply PPCA to individual modalities and PPCA, GPCCA, and MOFA to multi-modality data. For each method and cancer type, we perform Louvain clustering on the learned low-dimensional embeddings. Since no ground truth subject groups are available, we use distinction between survival time from phenotype data as a surrogate measure of cluster quality. NEMO and SNF are excluded from this study due to their inferior performance in previous analyses and their lack of low-dimensional embeddings, which prevents some downstream evaluations as detailed below.

\begin{figure}
    \centering
    \includegraphics[width=16cm]{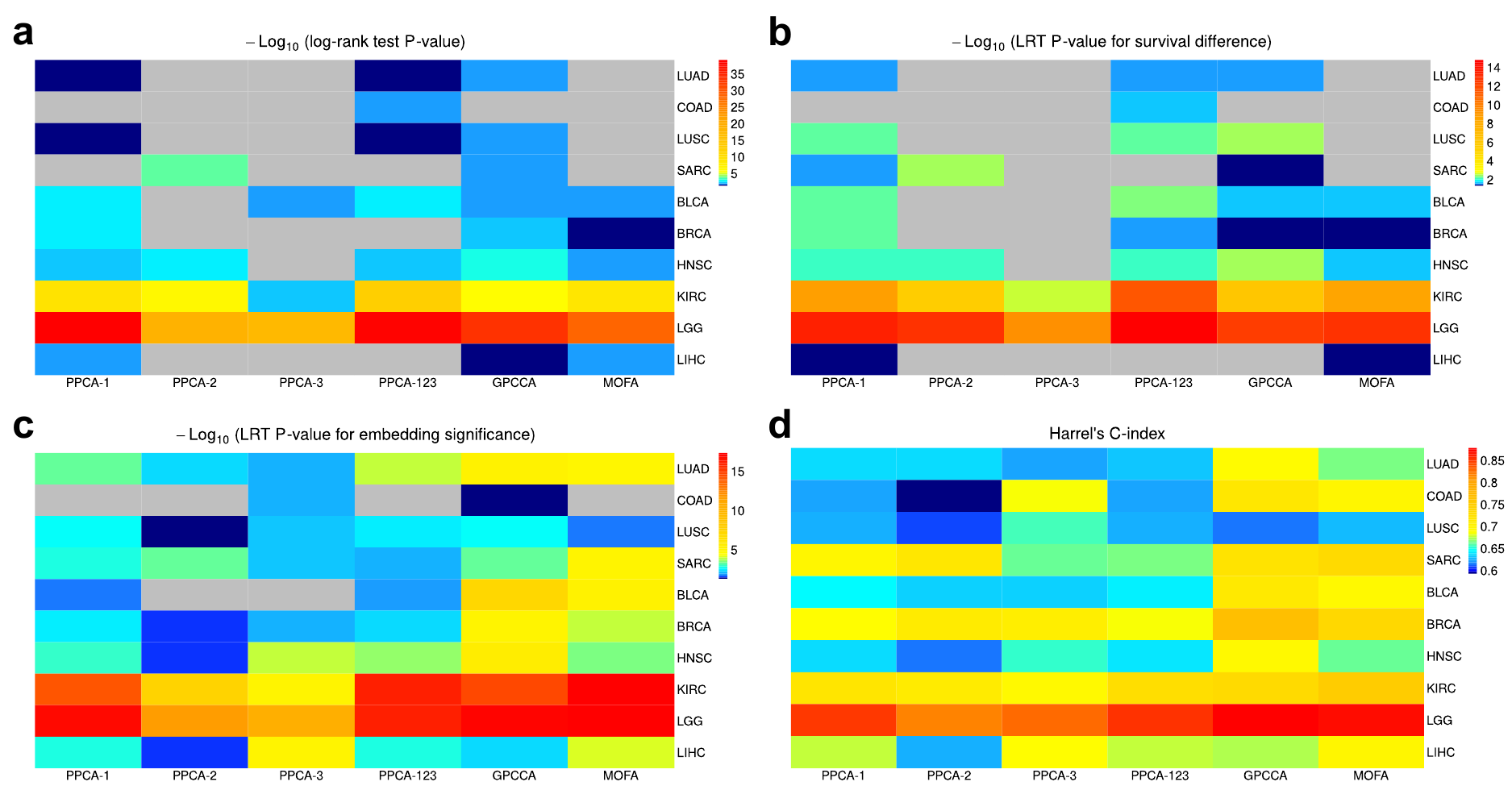}
    \caption{Method comparison based on the TCGA multi-omics data. \textbf{a.} Log-transformed adjusted $P$-values of log-rank test for comparing the survival time between identified clusters. \textbf{b.} Log-transformed adjusted $P$-values of LRT for comparing the survival time between identified clusters. \textbf{c.} Log-transformed adjusted $P$-values of LRT for testing the effect of data embeddings on survival prediction. \textbf{d.} C-index of the Cox regression model for survival prediction. Adjusted $P$-values above the threshold of 0.05 are shown as gray color.}
    \label{fig:TCGA - metrics}
\end{figure}

First, we perform two tests to evaluate whether there is significant difference in survival time between the identified subject clusters. The log-rank test is performed to directly evaluate survival differences between clusters, and likelihood-ratio test (LRT) based on the Cox regression model is used to evaluate survival differences after adjusting for two covariates: subject gender and age (at initial diagnosis). 
The $P$-values computed for all tested methods are adjusted by the Benjamini-Hochberg method.  Using the log-rank test, GPCCA identifies the most significant survival differences, reporting 9 out of 10 cancer types with $P$-values below the significance threshold of 0.05 (Figure \ref{fig:TCGA - metrics}a). PPCA applied to gene expression data (PPCA-1) leads to the second largest number of significant survival differences (8 out of 10). Using the LRT test, PPCA-1 identifies the most significant survival differences (9 out of 10), and GPCCA follows closely with 8 significant results (Figures \ref{fig:TCGA - metrics}b).
We also observe that, among the three modalities, gene expression levels generally contain more informative features for identifying differential survival groups than either DNA methylation or miRNA expression levels. However, for the sarcoma (SARC) dataset, methylation data alone is able to identify differential survival groups in the log-rank test, while gene expression data is not able to (Figure \ref{fig:TCGA - metrics}a). When applying GPCCA to the multi-omics SARC data, six clusters of subjects were identified, showing significant differences in survival rates (Supplementary Figure S12).  These results highlight GPCCA’s ability to integrate informative features from multiple modalities, circumventing the challenge of selecting a single modality for identifying survival groups.

Next, we apply Cox regression models to evaluate the predictive performance of the low-dimensional embeddings obtained by PPCA, GPCCA, and MOFA. Two approaches are considered. The first approach uses an LRT to assess whether adding the low-dimensional embeddings as predictors improves the model fit compared to using only subject gender and age. A higher significance indicates better predictive ability of the full model with embeddings, suggesting that the embeddings contain informative features related to subjects' survival time. The embeddings learned by GPCCA demonstrate statistically significant contributions to data fitting across all cancer types (Figure \ref{fig:TCGA - metrics}c). Furthermore, GPCCA's embeddings show greater significance than the embeddings learned by PPCA, whether applied to single-modality or multi-modality data.
The second approach uses Harrell's concordance index (C-index) \cite{harrell1982evaluating} to assess the model's ability to correctly rank subjects based on their predicted survival status. 
GPCCA and MOFA generally result in higher C-indices compared to the PPCA methods (Figure \ref{fig:TCGA - metrics}d). Notably, for lung adenocarcinoma (LUAD) and head and neck squamous cell carcinoma (HNSC) data, GPCCA demonstrates much higher C-indices than the other methods. These findings demonstrate the high informativeness of the low-dimensional projections learned by GPCCA, which improves survival prediction and leads to superior performance in Cox regression models.

\begin{table}[]
\centering\caption{
Comparison between informative and backgrounds gene sets across 10 TCGA cancer types.
``Total \# of Genes in OS'' represents the overlap between genes included in the OncoSearch database and those used in the multi-modal analysis.
For each background and informative gene set associated with a specific cancer type, the ``\% of genes in OS among all available'' is calculated as the number of overlapping genes divided by the total number of genes in OS, while the ``\% of genes in OS among all reported'' is the number of overlapping genes divided by the total number of genes in the given gene set.
}
\small{
\begin{tabular}{ |p{1.2cm}|p{2cm}|p{2.1cm}|p{2.1cm}|p{2.1cm}|p{2.1cm}| }
 \hline
 \multicolumn{1}{|c}{} & \multicolumn{1}{|c}{} & \multicolumn{2}{|c|}{Background gene set} & \multicolumn{2}{|c|}{Informative gene set} \\
 \hline
 Cancer type & Total \# of genes in OS & \# (or \%) of genes in OS among all available & \% of genes in OS among all reported & \# (or \%) of genes in OS among all available & \% of genes in OS among all reported \\
 \hline
 LUAD & $30$ & $5$ $(16.67\%)$ & $2.31\%$ & $12$ $(40.00\%)$ & $3.70\%$ \\
 COAD & $91$ & $11$ $(12.09\%)$ & $6.51\%$ & $43$ $(47.25\%)$ & $13.11\%$ \\
 LUSC & $4$ & $3$ $(75.00\%)$ & $0.50\%$ & $1$ $(25.00\%)$ & $0.53\%$ \\
 SARC & $81$ & $7$ $(8.64\%)$ & $4.43\%$ & $32$ $(39.51\%)$ & $9.82\%$ \\
 BLCA & $43$ & $4$ $(9.30\%)$ & $2.06\%$ & $32$ $(74.42\%)$ & $6.21\%$ \\
 BRCA & $113$ & $10$ $(8.85\%)$ & $6.49\%$ & $57$ $(50.44\%)$ & $11.78\%$ \\
 HNSC & $21$ & $2$ $(9.52\%)$ & $0.93\%$ & $15$ $(71.43\%)$ & $3.52\%$ \\
 KIRC & $4$ & $1$ $(25.00\%)$ & $0.22\%$ & $3$ $(75.00\%)$ & $0.81\%$ \\
 LGG  & $100$ & $17$ $(17.00\%)$ & $8.76\%$ & $58$ $(58.00\%)$ & $10.98\%$ \\
 LIHC & $107$ & $40$ $(37.38\%)$ & $9.30\%$ & $28$ $(26.17\%)$ & $15.30\%$ \\
 \hline
\end{tabular}
}
\label{tab:TCGA - OncoSearch}
\end{table}

To further investigate the latent factors identified by GPCCA, we focus on those that are statistically significant in the Cox models. Our goal is to compare gene features with varying loadings on these factors.
For each cancer type, we identify the top 100 genes from each significant factor based on the absolute values of their coefficients in the loading matrix. The union of these genes forms the ``informative gene set''. Conversely, we construct the ``background gene set'' by taking the complement of the top 100 genes within each factor and computing their intersection across all factors.
To compare these gene sets, we examine their overlap with genes in the OncoSearch database \cite{lee2014oncosearch}, which catalogs gene-cancer relationships by querying Medline abstracts. We find that the informative gene set exhibits greater overlap with OncoSearch-recorded genes than the background gene set (Table \ref{tab:TCGA - OncoSearch}), both in absolute numbers and relative proportions. This result supports the hypothesis that genes with higher loadings in GPCCA’s latent factors are more closely linked to cancer progression.
Additionally, we compare these gene sets with those in the OncoDB database \cite{tang2022oncodb}, which catalogs differentially expressed genes between normal and tumor samples. Across all cancer types, the informative gene set consistently exhibits smaller $P$-values than the background gene set (Supplementary Figure S13), which further confirms that the latent factors identified by GPCCA capture biologically relevant gene markers for different cancer types.

\section{Discussion} \label{sec:disc}

In this article, we introduce the GPCCA model for joint dimensionality reduction and multi-modal data integration. GPCCA learns low-dimensional embeddings of observations while preserving both shared and complementary information across modalities, leveraging cross-modal relationships to enhance downstream clustering analysis. It extends the classical CCA method in several ways. First, it formulates data integration within a probabilistic framework. Second, it generalizes to more than two modalities, enabling seamless integration of diverse data types. Third, it treats missing values as latent variables within the model, eliminating the need for separate imputation as a preprocessing step. We evaluate GPCCA through comprehensive simulation studies and two real-world applications. In the analysis of multi-view image data of handwritten digits, evaluations based on ground truth labels demonstrate GPCCA’s superior performance compared to alternative methods. In the application to multi-omics TCGA data, although ground truth subject subgroups are not available, our analysis shows that GPCCA effectively identifies patient groups with differential survival rates across cancer types. Furthermore, the learned latent factors reveal informative gene features closely associated with cancer, highlighting the model’s potential for biomedical discovery.

The development of GPCCA is motivated in part by the prevalence of missing values in real-world multi-modal data. Estimation of GPCCA is achieved through the proposed EM algorithm, which assumes data are missing at random (MAR). However, recognizing that real data often exhibit missing not at random (MNAR) patterns, we assess GPCCA’s robustness when the MAR assumption is violated. In both simulations (Case C) and real data applications, we specifically examine modality-wise missingness, a common scenario in biological and biomedical studies where the presence of a modality depends on latent, unobserved factors. Across multiple datasets with MNAR, GPCCA consistently demonstrates strong performance in identifying meaningful clusters based on the learned latent factors, suggesting its ability to handle complex missing patterns.

Despite the advantages discussed, we acknowledge several limitations of the current work that point to potential directions for future improvements and extensions. First, the computational cost of the EM algorithm increases with data dimensions, primarily due to the matrix inversion required for the error covariance matrix. While we employ block-wise inversion and other numerical techniques to mitigate this issue, runtime remains a challenge on high-dimensional data with missing values. To address this, we suggest that users consider a filtering step for modalities with too many features (e.g., the DNA methylation modality in TCGA data) by selecting the most variable features for integration. Furthermore, to enhance computational efficiency, we could explore incorporating low-rank approximations of the covariance matrix as an alternative approach.
Second, in our current applications of GPCCA to real data, we set the ridge regularization parameter 
$\lambda$ to 0.5, which has demonstrated good performance across a variety of simulation settings. However, in some applications, it may be beneficial to adaptively select $\lambda$ using grid search or other methods. Identifying a robust measure for this selection via cross-validation or other model selection techniques remains an important avenue for future work.
Third, GPCCA assumes that data follows a multivariate normal distribution. Although many datasets can be well-approximated by normal distributions after appropriate transformations, and our simulations have demonstrated GPCCA's robustness to heavy-tailed data (Case B), this assumption may limit its applicability to data modalities that significantly deviate from normality. A promising direction for future work is to extend GPCCA’s probabilistic framework to accommodate non-Gaussian distributions, such as binary or count data, to enhance its versatility.
In summary, while GPCCA demonstrates strong potential, these limitations also present exciting opportunities for further refinement, which could broaden its applicability and enhance its performance across a wide range of data types and use cases.

\section*{Declarations}

\subsection*{Ethics approval and consent to participate}
Not applicable.

\subsection*{Consent for publication
}
Not applicable.

\subsection*{Availability of data and material}
The GPCCA model has been implemented as an R package, which is freely available from its Github repository \url{https://github.com/Kaversoniano/GPCCA}. The real datasets analyzed during the current study are both available online. The multi-view image data can be downloaded from \url{https://archive.ics.uci.edu/dataset/72/multiple+features}. The multi-omics TCGA data can be downloaded from\url{https://xenabrowser.net/datapages/}.

\subsection*{Competing interests
}
The authors have no competing interests.

\subsection*{Funding}
This work was partially supported by NIH R35GM142702 (to WVL).

\subsection*{Authors' contributions}
WVL conceptualized and supervised the study. TY and WVL developed the methodology. TY implemented the algorithm, conducted the analyses, and designed the software package. Both authors contributed to drafting and revising the manuscript. Both authors read and approved the final manuscript.

\subsection*{Acknowledgment}
The authors would like to thank Dr. Esra Kurum, Dr. Weixin Yao, and Dr. Adam Godzik at University of California, Riverside (UCR) and other members of the Vivian Li Lab for their insightful suggestions on this work.
The authors acknowledge the HPC Center at UCR and NSF-MRI grant 2215705 for the computing resources made available for conducting the research reported in this paper.

\bibliographystyle{unsrt}
\bibliography{main}

\begin{thebibliography}{10}

\bibitem{guo2019canonical}
Chenfeng Guo and Dongrui Wu.
\newblock Canonical correlation analysis (cca) based multi-view learning: An
  overview.
\newblock {\em arXiv preprint arXiv:1907.01693}, 2019.

\bibitem{kalamkar2023multimodal}
Shrida Kalamkar et~al.
\newblock Multimodal image fusion: A systematic review.
\newblock {\em Decision Analytics Journal}, page 100327, 2023.

\bibitem{cancer2008comprehensive}
Cancer Genome Atlas Research Network~Tissue source~sites: Duke University
  Medical School McLendon Roger 1 Friedman Allan 2 Bigner Darrell~1, Emory
  University Van Meir Erwin G. 3 4 5 Brat Daniel J. 5 6 M. Mastrogianakis Gena
  3 Olson Jeffrey J. 3~4 5, Henry Ford Hospital Mikkelsen Tom 7 Lehman~Norman
  8, MD~Anderson Cancer Center Aldape Ken 9 Alfred Yung WK 10 Bogler~Oliver 11,
  University of~California San Francisco VandenBerg Scott 12 Berger Mitchel 13
  Prados Michael~13, et~al.
\newblock Comprehensive genomic characterization defines human glioblastoma
  genes and core pathways.
\newblock {\em Nature}, 455(7216):1061--1068, 2008.

\bibitem{hao2021integrated}
Yuhan Hao, Stephanie Hao, Erica Andersen-Nissen, William~M Mauck, Shiwei Zheng,
  Andrew Butler, Maddie~J Lee, Aaron~J Wilk, Charlotte Darby, Michael Zager,
  et~al.
\newblock Integrated analysis of multimodal single-cell data.
\newblock {\em Cell}, 184(13):3573--3587, 2021.

\bibitem{rappoport2018multi}
Nimrod Rappoport and Ron Shamir.
\newblock Multi-omic and multi-view clustering algorithms: review and cancer
  benchmark.
\newblock {\em Nucleic acids research}, 46(20):10546--10562, 2018.

\bibitem{leng2022benchmark}
Dongjin Leng, Linyi Zheng, Yuqi Wen, Yunhao Zhang, Lianlian Wu, Jing Wang,
  Meihong Wang, Zhongnan Zhang, Song He, and Xiaochen Bo.
\newblock A benchmark study of deep learning-based multi-omics data fusion
  methods for cancer.
\newblock {\em Genome biology}, 23(1):171, 2022.

\bibitem{flores2023missing}
Javier~E Flores, Daniel~M Claborne, Zachary~D Weller, Bobbie-Jo~M
  Webb-Robertson, Katrina~M Waters, and Lisa~M Bramer.
\newblock Missing data in multi-omics integration: Recent advances through
  artificial intelligence.
\newblock {\em Frontiers in Artificial Intelligence}, 6:1098308, 2023.

\bibitem{mackiewicz1993principal}
Andrzej Ma{\'c}kiewicz and Waldemar Ratajczak.
\newblock Principal components analysis (pca).
\newblock {\em Computers \& Geosciences}, 19(3):303--342, 1993.

\bibitem{nguyen2017novel}
Tin Nguyen, Rebecca Tagett, Diana Diaz, and Sorin Draghici.
\newblock A novel approach for data integration and disease subtyping.
\newblock {\em Genome research}, 27(12):2025--2039, 2017.

\bibitem{wang2014similarity}
Bo~Wang, Aziz~M Mezlini, Feyyaz Demir, Marc Fiume, Zhuowen Tu, Michael Brudno,
  Benjamin Haibe-Kains, and Anna Goldenberg.
\newblock Similarity network fusion for aggregating data types on a genomic
  scale.
\newblock {\em Nature methods}, 11(3):333--337, 2014.

\bibitem{rappoport2019nemo}
Nimrod Rappoport and Ron Shamir.
\newblock Nemo: cancer subtyping by integration of partial multi-omic data.
\newblock {\em Bioinformatics}, 35(18):3348--3356, 2019.

\bibitem{brunet2004metagenes}
Jean-Philippe Brunet, Pablo Tamayo, Todd~R Golub, and Jill~P Mesirov.
\newblock Metagenes and molecular pattern discovery using matrix factorization.
\newblock {\em Proceedings of the national academy of sciences},
  101(12):4164--4169, 2004.

\bibitem{liu2013multi}
Jialu Liu, Chi Wang, Jing Gao, and Jiawei Han.
\newblock Multi-view clustering via joint nonnegative matrix factorization.
\newblock In {\em Proceedings of the 2013 SIAM international conference on data
  mining}, pages 252--260. SIAM, 2013.

\bibitem{hotelling1992relations}
Harold Hotelling.
\newblock Relations between two sets of variates.
\newblock In {\em Breakthroughs in statistics: methodology and distribution},
  pages 162--190. Springer, 1992.

\bibitem{witten2009extensions}
Daniela~M Witten and Robert~J Tibshirani.
\newblock Extensions of sparse canonical correlation analysis with applications
  to genomic data.
\newblock {\em Statistical applications in genetics and molecular biology},
  8(1), 2009.

\bibitem{lin2023quantifying}
Kevin~Z Lin and Nancy~R Zhang.
\newblock Quantifying common and distinct information in single-cell multimodal
  data with tilted canonical correlation analysis.
\newblock {\em Proceedings of the National Academy of Sciences},
  120(32):e2303647120, 2023.

\bibitem{pedersen2017missing}
Alma~B Pedersen, Ellen~M Mikkelsen, Deirdre Cronin-Fenton, Nickolaj~R
  Kristensen, Tra~My Pham, Lars Pedersen, and Irene Petersen.
\newblock Missing data and multiple imputation in clinical epidemiological
  research.
\newblock {\em Clinical epidemiology}, pages 157--166, 2017.

\bibitem{tipping1999probabilistic}
Michael~E Tipping and Christopher~M Bishop.
\newblock Probabilistic principal component analysis.
\newblock {\em Journal of the Royal Statistical Society Series B: Statistical
  Methodology}, 61(3):611--622, 1999.

\bibitem{severson2017principal}
Kristen~A Severson, Mark~C Molaro, and Richard~D Braatz.
\newblock Principal component analysis of process datasets with missing values.
\newblock {\em Processes}, 5(3):38, 2017.

\bibitem{yu2010probabilistic}
Lingbo Yu, Robert~R Snapp, Teresa Ruiz, and Michael Radermacher.
\newblock Probabilistic principal component analysis with expectation
  maximization (ppca-em) facilitates volume classification and estimates the
  missing data.
\newblock {\em Journal of structural biology}, 171(1):18--30, 2010.

\bibitem{van2012generalized}
Michel Van~de Velden and Yoshio Takane.
\newblock Generalized canonical correlation analysis with missing values.
\newblock {\em Computational Statistics}, 27:551--571, 2012.

\bibitem{argelaguet2018multi}
Ricard Argelaguet, Britta Velten, Damien Arnol, Sascha Dietrich, Thorsten Zenz,
  John~C Marioni, Florian Buettner, Wolfgang Huber, and Oliver Stegle.
\newblock Multi-omics factor analysis—a framework for unsupervised
  integration of multi-omics data sets.
\newblock {\em Molecular systems biology}, 14(6):e8124, 2018.

\bibitem{bach2005probabilistic}
Francis~R Bach and Michael~I Jordan.
\newblock A probabilistic interpretation of canonical correlation analysis.
\newblock 2005.

\bibitem{dempster1977maximum}
Arthur~P Dempster, Nan~M Laird, and Donald~B Rubin.
\newblock Maximum likelihood from incomplete data via the em algorithm.
\newblock {\em Journal of the royal statistical society: series B
  (methodological)}, 39(1):1--22, 1977.

\bibitem{woodbury1950inverting}
Max~A Woodbury.
\newblock {\em Inverting modified matrices}.
\newblock Department of Statistics, Princeton University, 1950.

\bibitem{warton2008penalized}
David~I Warton.
\newblock Penalized normal likelihood and ridge regularization of correlation
  and covariance matrices.
\newblock {\em Journal of the American Statistical Association},
  103(481):340--349, 2008.

\bibitem{blondel2008fast}
Vincent~D Blondel, Jean-Loup Guillaume, Renaud Lambiotte, and Etienne Lefebvre.
\newblock Fast unfolding of communities in large networks.
\newblock {\em Journal of statistical mechanics: theory and experiment},
  2008(10):P10008, 2008.

\bibitem{roweis1997algorithms}
Sam Roweis.
\newblock Em algorithms for pca and spca.
\newblock {\em Advances in neural information processing systems}, 10, 1997.

\bibitem{ng2001spectral}
Andrew Ng, Michael Jordan, and Yair Weiss.
\newblock On spectral clustering: Analysis and an algorithm.
\newblock {\em Advances in neural information processing systems}, 14, 2001.

\bibitem{yeung2001details}
Ka~Yee Yeung and Walter~L Ruzzo.
\newblock Details of the adjusted rand index and clustering algorithms,
  supplement to the paper an empirical study on principal component analysis
  for clustering gene expression data.
\newblock {\em Bioinformatics}, 17(9):763--774, 2001.

\bibitem{stacklies2007pcamethods}
Wolfram Stacklies, Henning Redestig, Matthias Scholz, Dirk Walther, and Joachim
  Selbig.
\newblock pcamethods—a bioconductor package providing pca methods for
  incomplete data.
\newblock {\em Bioinformatics}, 23(9):1164--1167, 2007.

\bibitem{zhuang2022findpc}
Haotian Zhuang, Huimin Wang, and Zhicheng Ji.
\newblock findpc: An r package to automatically select the number of principal
  components in single-cell analysis.
\newblock {\em Bioinformatics}, 38(10):2949--2951, 2022.

\bibitem{multiple_features_72}
Robert Duin.
\newblock {Multiple Features}.
\newblock UCI Machine Learning Repository, 1998.
\newblock {DOI}: https://doi.org/10.24432/C5HC70.

\bibitem{harrell1982evaluating}
Frank~E Harrell, Robert~M Califf, David~B Pryor, Kerry~L Lee, and Robert~A
  Rosati.
\newblock Evaluating the yield of medical tests.
\newblock {\em Jama}, 247(18):2543--2546, 1982.

\bibitem{lee2014oncosearch}
Hee-Jin Lee, Tien~Cuong Dang, Hyunju Lee, and Jong~C Park.
\newblock Oncosearch: cancer gene search engine with literature evidence.
\newblock {\em Nucleic acids research}, 42(W1):W416--W421, 2014.

\bibitem{tang2022oncodb}
Gongyu Tang, Minsu Cho, and Xiaowei Wang.
\newblock Oncodb: an interactive online database for analysis of gene
  expression and viral infection in cancer.
\newblock {\em Nucleic Acids Research}, 50(D1):D1334--D1339, 2022.

\end{thebibliography}

\end{document}